\documentclass{svjour3}

\usepackage{xspace}

\usepackage{import}
\usepackage{soul}
\usepackage[utf8]{inputenc}
\usepackage{amsmath}
\usepackage{amssymb}
\usepackage{booktabs}
\usepackage{algorithm}
\usepackage[noend]{algpseudocode}
\usepackage{bbm}
\usepackage{calc}
\usepackage{textcomp}
\usepackage{hyperref}

\usepackage{color}

\usepackage{mathtools}
\usepackage{cite}

\ifdefined\releaseversion
    \newcommand{\andrea}[1]{}
    \newcommand{\luca}[1]{}
    \newcommand{\fausto}[1]{}
\else
    \newcommand{\andrea}[1]{{\bf \textcolor{blue}{Andrea: #1}}}
    \newcommand{\luca}[1]{{\bf \textcolor{cyan}{Luca: #1}}}
    \newcommand{\fausto}[1]{{\bf \textcolor{red}{Fausto: #1}}}
\fi

\usepackage{standalone}
\usepackage{tikz}

\usetikzlibrary{calc}
\usetikzlibrary{math}
\pgfdeclarelayer{background}
\usetikzlibrary{positioning, patterns, fit}
\pgfsetlayers{background,main}

\tikzset{
    pics/linkedcouple/.style  args={(#1) (#2)}{code={
        \node[pic actions,fit = (#1)] (temp1) {} ;
        \node[pic actions,fit = (#2)] (temp2) {} ;
        \draw (temp1) -- (temp2);
    }},
    matchstyle/.style={draw, green!60!black, line width=.5mm, inner sep =1mm, rounded corners=2mm},
    nomatchstyle/.style={matchstyle, red!40!gray},
    dbgpoint/.style={circle,inner sep=0pt,minimum size=2pt,fill=red},
	keypoint/.style={inner sep=0pt,minimum size=0pt},
    picframe/.style={inner sep=0pt,minimum size=6mm},
	enc/.style={draw,rounded corners=2mm, inner sep= 4mm, line width=0.3mm},
	worm/.style={draw,dashed, rounded corners=2mm, draw opacity=0.5, line width=0.6mm, inner sep =4mm},
    missingimg/.style={fill=gray!10!white, text=black,text width =1cm,  minimum size = 18mm, path picture={
        \draw[gray!30!white]
               (path picture bounding box.south east) -- (path picture bounding box.north west) 
               (path picture bounding box.south west) -- (path picture bounding box.north east);
         \node[anchor=center, gray!60!white, text width=1cm] at (path picture bounding box.center) {missing image} ;
          } 
     }
}

\def\argmax{\operatornamewithlimits{argmax}}

\def \calK{{\cal K}}
\def \calM{{\cal M}}

\newcommand{\defas}{\ensuremath{\stackrel{\text{\tiny def}}{=}}\xspace}

\def \genus{\texttt{Genus}\xspace}
\def \genusobj{\texttt{genusObj}\xspace}
\def \genusfun{\texttt{genusOf}\xspace}
\def \same{\texttt{sameGenus}\xspace}
\def \D{\texttt{D}\xspace}

\def \differentia{\texttt{Differentia}\xspace}
\def \different{\texttt{Different}\xspace}

\def \objnn{\ensuremath{O}\xspace}

\newcommand{\enc}{\ensuremath{E}\xspace}

\newcommand{\obj}{\ensuremath{O}\xspace}

\newcommand{\encS}{\ensuremath{E}_{S}\xspace}
\newcommand{\objS}{\ensuremath{O}_{S}\xspace}
 \newcommand{\foSi}[1]{\ensuremath{f}_{#1}\xspace}

 \newcommand{\foS}[2]{\ensuremath{f}^{#1}_{#2}\xspace}

\newcommand{\voSi}[1]{\ensuremath{v}_{#1}\xspace}

\newcommand{\voS}[2]{\ensuremath{v}^{#1}_{#2}\xspace}

\newcommand{\EoS}[2]{\ensuremath{E}^{#1}_{#2}\xspace}

\newcommand{\objSi}[1]{\ensuremath{O}_{#1}\xspace}

\newcommand{\objSii}[2]{\ensuremath{O}^{#1}_{#2}\xspace}

\renewcommand{\hl}[1]{#1}

\title{Towards Visual Semantics}

 \author{ Fausto Giunchiglia  \and Luca Erculiani\and Andrea Passerini 
 }
 \institute{
 Fausto Giunchiglia \and Luca Erculiani\and Andrea Passerini \at University of Trento, Italy \\ 
\email{{\it name.surname}@unitn.it} \\
Corresponding Author: Fausto Giunchiglia
 }

\begin{document}
\maketitle

\begin{abstract} 
\hl{\textit{Lexical Semantics} is concerned with how words encode mental representations of the world, i.e., \textit{concepts}. We call this type of concepts, \textit{classification concepts}. In this paper, we  focus on \textit{Visual Semantics}, namely on how humans build  concepts representing what they perceive visually. We call this second type of concepts, \textit{substance concepts}. As shown in the paper, these two types of concepts are different and, furthermore, the mapping between them is many-to-many. In this paper we provide a theory and an algorithm for how to build substance concepts which are in a one-to-one correspondence with classifications concepts, thus paving the way
to the seamless integration between natural language descriptions and visual perception. }
This work builds upon three main intuitions: (i) substance concepts are modeled as \textit{visual objects}, namely sequences of similar frames, as perceived in multiple \textit{encounters}; (ii) substance concepts are organized into a \textit{visual subsumption hierarchy} based on the notions of \genus and \differentia; (iii) the human feedback is exploited \textit{not} to name objects, but, rather, to align the hierarchy of substance concepts with that of classification concepts. The learning algorithm is implemented for the base case of a hierarchy of depth two. The experiments, though preliminary, show that the algorithm manages to acquire the notions of \genus and \differentia with reasonable accuracy, this despite seeing
a small number of examples and receiving supervision on a fraction of them.
\end{abstract}

\section{Introduction}
\label{sec-intro}

\hl{The Oxford Research Encyclopedia defines \textit{Lexical Semantics} as the study of \textit{word meanings}, i.e., \textit{concepts} }\cite{lalumera2010concepts},\hl{ where concepts are assumed to be  constructed by humans through language.  
In the same line of thinking, this research focuses on \textit{Visual Semantics}, namely on how humans build concepts when using vision to perceive the world. 
The key assumption is that these two types of concepts are different and that, furthermore, they stand in a many-to-many relation (see Section} \ref{sec-lsm} for the details).\footnote{\hl{This assumption is consistent with the fact that the two activities of speaking and seeing involve different parts of the human brain } \cite{martin2001semantic}.} Following the terminology from \cite{giunchiglia2016fois}, we call the first type of concepts, \textit{classification concepts}, and the latter type, \textit{substance concepts}.\footnote{This terminology is motivated by the fundamentally different \textit{function} that these concepts have. 
In fact, while the substance concepts are used to \textit{represent substances} as they are perceived, the latter are used to \textit{describe what is  perceived, i.e., substance concepts}.
This idea of seeing concepts as (biological) \textit{functions} is based on the work in the field of \textit{Teleosemantics}, sometimes called \textit{Biosemantics} \cite{macdonald2006teleosemantics}, and in particular on the work by the philosopher R. Millikan \cite{millikan1984language,millikan1998more,millikan2000clear, millikan2005language}.
}
%
\hl{Our goal in this paper is to provide a theory and an algorithm for how to build substance concepts which are in a one-to-one correspondence with classifications concepts, thus paving the way
to the seamless integration between natural language descriptions and visual perception. 
Among other things, the solution we propose allows to deal with the so-called \textit{Semantic Gap Problem} (SGP)} \cite{2010smeulders}. \hl{The SGP, originally identified in 2010 ans still largely unsolved, arises from the fact that, in general, there is a misalignment between what Computer Vision systems perceive from media and the words that humans use to describe the same sources. }
We articulate the problem we deal with is as follows.

\hl{\textit{Suppose that a person and a machine, e.g., a pair of smart glasses, are such that they see the same parts of the world under the same visual conditions.  
Suppose that the person has a full repertoire of words which allow her to describe what she sees according to her current point of view. 
Suppose, furthermore, that the machine starts from scratch without any prior knowledge of the world and of how to name whatever it perceives. How can we build an algorithm which, by suitably asking the human,  will learn how to recognize and name whatever it sees in the same way as its reference user?}}

A meaningful metaphor for this problem is that of a mother who is teaching her baby child how to name things using her own words in her own spoken language. The work in  \cite{giunchiglia2016fois} provides an extensive description of the complicacies related to this problem, mainly related to the many-to-many relation existing between substance and classification concepts. Further complications come from the fact that, based on the definition above, the learning algorithm needs to satisfy the following further requirements:
 
\begin{itemize}
\item
it must be generic, in that it should make no assumptions about the input objects; 
\item 
it must learn new objects never seen before as well as
novel features, never seen before, of previously seen objects; 
\item 
it must learn from a small number of examples, starting from no examples.
\end{itemize}

\hl{The proposed Knowledge Representation (KR) solution is articulated in terms of a set of novel definitions of some basic notions, most importantly that of \textit{object}. }
The theory proceeds as follows. 

\begin{itemize}
    \item We model \textit{objects} as \textit{substance concepts}, that we model as sets of \textit{visual objects}, i.e., sequences of similar frames, as perceived in multiple events called \textit{encounters}. Visual objects are stored in a \textit{cumulative memory} $\calM$ of all the times they were previously perceived. 

\item
Substance concepts are organized into a \textit{(visual) subsumption hierarchy} which is learned based on the notions  of \genus and \differentia. These two notions \textit{mutatis mutandis}, replicate the notions with the same name that, in \textit{Lexical Semantics}, are used to build \textit{subsumption hierarchies} of word meanings \cite{miller1990introduction,giunchiglia2017understanding}. 

\item The visual hierarchy is learned autonomously by the algorithm; the user feedback makes sure that the hierarchy built by the machine matches her own linguistic organization of objects. In other words, the user feedback is the means by which the hierarchy of substance concepts is transformed into a hierarchy of classification concepts. The key observation here is that the user feedback is provided not in terms of object names, as it is usually the case, but in terms of the two properties of \genus and \differentia.
\end{itemize}

The paper is organized as follows. First, we introduce objects as classification concepts, as they are used in natural language and organized in Lexical Semantics hierarchies (Section \ref{sec-lsm}). This section provides also an analysis of why the very definition of classification concepts makes them unsuitable for visual object recognition.
Then we define substance concepts as sets of visual objects (Section \ref{sec-substance}).  Then, in Section  \ref{sec3-createsc}, we provide the main algorithm by which substance concepts are built, while, in Section \ref{sec-visual}, we describe how a hierarchy of substance concepts is built which is aligned with that of classification concepts. In this section we also provide the two basic notions of \genus and \differentia which are used to build the hierarchy. The algorithm for
object learning is described in Section \ref{sec-algorithm}. This algorithm has been developed for the base case of  hierarchies of depth two. The extension to hierarchies of any level is left to the future work. The algorithm in evaluated in Section \ref{sec-evaluation}. Finally, the paper ends with the related work (Section \ref{sec-related}) and the conclusions (Section \ref{sec-conclusion}).

\section{Objects as Classification Concepts}
\label{sec-lsm}

Objects are usually named using nouns. In Lexical Semantics the meaning of nouns is provided via intensional definitions articulated in terms of \genus and \differentia \cite{miller1990introduction}, following an approach first introduced by Aristotle \cite{Aristotle}. Let us consider for instance the following two definitions:
\begin{itemize}
    \item a \textit{triangle} is a \textit{plane figure} with \textit{three straight bounding sides};
    \item a \textit{quadrilateral} is a \textit{plane figure} with \textit{four straight bounding sides}.
\end{itemize}
\noindent
In these two definitions we can identify three main components:
\begin{itemize}
    \item \genus: some previously defined set of properties which is shared across distinct objects, e.g., the property of  \textit{being a plane figure};
    \item \genusobj (also called \genusobj object): a certain representative object which satisfies the \genus property, e.g., the object \textit{plane figure}. The set of objects satisfying the \textit{Genus} properties are said to have that (same) \genusobj;
    \item \differentia: A selected novel set of properties, different from the \genus properties, which are used to differentiate among objects with the same \genusobj, e.g., the properties \textit{having three straight bounding sides} and \textit{having fours straight bounding sides}. These two properties define, respectively, triangles and quadrilaterals as distinct objects with the same \genusobj.
\end{itemize}
\noindent \genus and \differentia satisfy the following four constraints:

\begin{itemize}

\item \textit{Role 1 of} \genus: if two 
 objects have different \genusobj, then they are (said to be) \textit{different}. For instance, a pyramid is not a plane figure and, therefore, is different from a triangle. 
 
\item \textit{Role 2 of} \genus: The viceversa of Role 1 is not true, namely we may have different objects with the same \genusobj.
For instance, a quadrilateral and a triangle are both plane figures but they are not the same object.

\item \textit{Role 1 of} \differentia: Two objects with the same \genusobj, but different from the \genusobj,  are (said to be) the \textit{same} object if and only if the \differentia properties do not hold of the two objects.
Thus, for instance, two objects with the same \genusobj and with a different \differentia, e.g., 
a triangle and a quadrilateral, are different despite being both a plane figure. Dually, two objects with the same \genusobj and the same \differentia, e.g., two triangles, are the same object (relatively to the current selection of \genus and \differentia).

\item \textit{Role 2 of} \differentia: a \genusobj and an object with that \genusobj are different when the latter is characterized by a set of properties, i.e., its \differentia, that the \genusobj does not have. 
Thus for instance a  triangle is not the same as a plane figure, as it is just one of the many possible plane figures,
e.g., triangles, quadrilaterals which share the same \genusobj.
\end{itemize}

A first observation about the definitions above is that, when we say that two objects are the same object, we only mean  that they satisfy the same  \genus and the same \differentia. It does not necessarily mean that they are two occurrences of the same object.
Thus, for instance, a right triangle and a equilateral triangle are considered as being the same object, when compared with quadrilaterals, in that they have the same number of sides. At the same time they are considered as different objects when the \differentia is taken to be  the size of their angles. This observation has two immediate consequences. The first is that the process above can be iterated at any level of detail, thus creating hierarchies of any level of depth.
It is a fact that, in lexical semantics, the meaning of nouns is organized as a hierarchy of increasing specificity,
each layer being characterized by a new \genus and a new \differentia. In this hierarchy, an object with a certain \genus 
is  a child of its \genusobj. As a consequence, a hierarchy of depth $n$ can be seen as the recursive juxtaposition of $(n-1)$ hierarchies of depth $2$, where the \genusobj of the depth $2$ hierarchy one level down is one of the children of the \genusobj one level above. 
The root of this hierarchy is usually called \textit{thing} \cite{miller1990introduction,giunchiglia2017understanding}. 
The second is that this process of progressive differentiation allows to split the set of objects under consideration into progressively smaller and smaller sets, based on the selected set of properties.

A second observation is that the above definitions are given in natural language and are meant to make precise the meaning of words. These linguistic definitions are designed to generate what we call \textit{classification concepts}, namely concepts which are amenable for classification \cite{2019-Fumagalli2}. And in fact the very existence of lexical semantics hierarchies provides evidence of their suitability for this task.
This type of definitions is well grounded in the everyday practice, in particular when used to name and describe things, for instance during interactions among humans. 
However they do not work as well while one is in the middle of the recognition process, namely while she is trying to identify the object she is looking at. How many times were you able to recognize someone or something based only on a natural language description, without the help of a photo or anything which could point to specific spatial properties?

\begin{figure}[t]
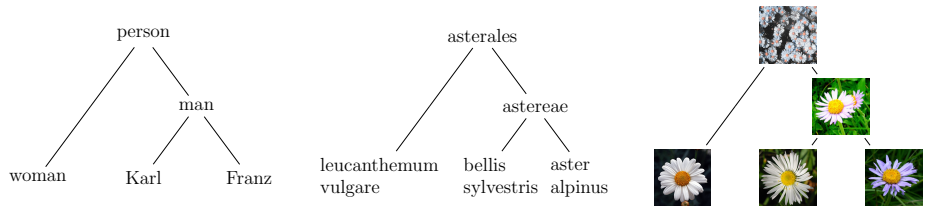

    \centering
    \includestandalone[width=1\textwidth]{apples}
    \caption{\label{fig:visual_objects}(Left): a classification concept hierarchy; (Center): a classification concept hierarchy for daisies; (Right:) the center hierarchy where words are substituted with images representing the corresponding daisies.}
 \label{fig:lexsem}    
\end{figure}

Let us clarify this observation with an example. Assume you see at a certain distance two \textit{things} moving towards you. Initially you will not recognize what these things are but, when they are close enough, you will be able to recognize two \textit{person}s, seen from the back. The day after, you see again two persons, which may or may not be those recognized the day before: hard to say, they did not come close enough. In any case, this second time these two persons get close enough for you to finally recognize your friends \textit{Karl} and \textit{Frank}. What allowed you to distinguish  \textit{Karl} from \textit{Frank} is that the former has white hair while the latter has black air and mustaches. Later on, walking towards you,  you will recognize a \textit{woman}. You will have been able to recognize her as a \textit{person} different from the two previous \textit{men} because she has long hair and a skirt. 
Of course you will know the terms you have used to describe what you will have seen, i.e., \textit{person}, \textit{man}, \textit{woman}, \textit{Karl} and \textit{Frank}, as someone will have taught them to you, for instance during your childhood.
In KR, the simple scene described above can be formalized by saying that \textit{Karl} and \textit{Frank} are \textit{instances}, while \textit{person}, \textit{man} and \textit{woman} are \textit{(classification) concepts} 
and by stating the following facts: \textit{man(Karl)} (to be read as \textit{Karl is a man}),  \textit{man(Frank)}, \textit{man $\sqsubseteq$ person} (to be read as \textit{man is subsumed by person}) and \textit{woman $\sqsubseteq$ person}, the latter two facts stating that all men and all women are persons. The resulting hierarchy, as formally defined via the logical subsumption symbol $\sqsubseteq$, is provided in Figure \ref{fig:lexsem} (first left) where the classification concepts there represented are defined, for instance, as (partial quote from \cite{miller1990introduction})
\begin{itemize}
    \item \textit{person}: individual, someone, somebody;
    \item \textit{woman}: an adult female person, as opposed to man;
    \item \textit{man}: an adult male person, as opposed to woman;
    \item \textit{Karl}: an instance of a man;
    \item \textit{Franz}: an instance of a man.
\end{itemize}
\noindent
Notice how the above definitions and the properties they involve (e.g., being adult, male or female, being an instance) are completely unrelated to the process by which recognition was carried out, which was in terms of a continual analysis of visual information, at increasing levels of precision. 

The previous example is representative of the situation where the observer has complete knowledge of the objects being perceived and the partiality of information is caused by some contextual factors. Consider now the hierarchy of classification concepts in the center of Figure \ref{fig:lexsem}, which names and classifies daisies, whose images are in the corresponding place in the hierarchy in the right of Figure \ref{fig:lexsem}. A possible lexical semantics definition of these daisies is as follows:
\begin{itemize}
    \item \textit{asterales}: an order of flowering plants containing eleven families, the most notable being asteraceae (known for composite flowers made of florets);
    \item \textit{leucanthenum vulgare}: flower native to Europe and the temperate regions of Asia, commonly referred as marguerite;
    \item \textit{astereae}: a tribe of plants, commonly found in temperate regions of the world, also called daisy or sunflower family;
    \item \textit{bellis sylvestris}: Southern daisy, perennial plant native to central and northern Europe;
    \item \textit{aster alpinus}: blue alpine daisy, plant commonly found in the mountains in Europe.
\end{itemize}
Most readers, in particular those who are not florists, even if coming to know about the hierarchy above, e.g., because being described it, will be unable to recognize the various types of daisy. As a consequence they will not be able to build it starting from images (e.g., the ones on the right in Figure \ref{fig:lexsem}), simply because they will not be able to recognize the features which allow to distinguish among the various types of daisy. Most likely, in many cases, the hierarchy will be collapsed to a single node while, in others, the light purple daisy will be separated from the others, just because of its colour.

In general, classification concepts do not seem well suited for the process of object recognition. This despite the fact that it is common practice to use them in supervised machine learning, where the user feedback is, often if not always, provided in terms of words whose meaning is defined via lexical semantics hierarchies. Evidence of this difficulty is provided by the SGP, whose original formulation describes it as (quote from
 \cite{2010smeulders}) ``\textit{... the lack of coincidence between the information that one can extract from the visual data and the interpretation that the same data have for a user in a given situation.}". 
 The main motivation for the SGP seems to be that 
 classification concepts model objects as \textit{endurants}, i.e., as being always wholly present,  at any given moment in time,  with their proper parts being present in a certain spatial configuration and satisfying certain properties (e.g., color, shape, position, activity) \cite{gangemi2002sweetening}. Typical examples of endurants are all the physical objects, e.g., those mentioned above. Of course, the spatial configuration may change, or the object might not be accessible visually (as in the first example above), or the observer might not be able to discriminate some of its relevant properties (as in the daisies example above), but this has no impact on how classification concepts are defined.

Classification concepts, while serving well the purpose of describing what was previously perceived, are largely unrelated to the process by which the objects are perceived and, in particular, to the fact that their perception is constructed \textit{incrementally}, via a set of \textit{partial views} which progressively enrich what is visually known. To this extent, notice how \textit{person}, \textit{man}, and \textit{Karl} are correctly represented in Figure \ref{fig:lexsem} as three different classification concepts. However in the little story above, these three classification concepts actually describe the same piece of reality, seen at different times, at different levels of detail and from different points of view.

\section{Objects as substance concepts}
\label{sec-substance}

The key intuition underlying the work described here is to model objects are \textit{perdurants}, where, quoting from  \cite{gangemi2002sweetening} \textit{... perdurants ... just extend in time by accumulating different temporal parts, so that,
at any time they are present, they are only partially present, in the sense that some of
their proper temporal parts (e.g., their previous or future phases) may be not present.} Typical examples of perdurants are events and activities. Taking an object as a perdurant amounts to saying that we never have a full (visual) picture of the object but that its visual representation is built progressively, in time. Notice how this is exactly what happens in our everyday life. We call \textit{substance concepts} the representation of objects as perdurants.

The starting point in the definition of substance concepts is the crucial distinction between what we perceive as being in the real world, that we call \textit{substances} and their corresponding mental representations, i.e., their substance concepts. Following R. Millikan, we take substances as  those things (quote from \cite{millikan2000clear}),  ``\textit{... about which you can learn from one encounter something of what to expect on other encounters, where this is no accident but the result of a real connection ...}". 
\cite{giunchiglia2016fois} provides a detailed discussion of what substances are and of how they generate substance concepts in the mind of observers, based on the work on \textit{Teleosemantics} 
 \cite{macdonald2006teleosemantics}, and in particular on the work by Ruth Millikan \cite{millikan1984language,miller1990introduction,millikan1989biosemantics,millikan2004varieties,millikan1998more,millikan2000clear}. In the following, substances should be intuitively thought as those things which, when perceived in the most detail, will generate the perception of individuals, e.g., \textit{Karl}, \textit{my cat}, but that, under different conditions, will generate more generic or even very different substance concepts, e.g., \textit{a moving object}, \textit{an animal}.
 The key observation is that, while substances are crucial in our informal understanding of perception in that they allow us to focus on the process of how objects are perceived, 
they play no role in the formal model that we define below. 
With this in mind in the following: (i) 
we avoid defining what a substance is 
(no such definition could be meaningfully grounded in human experience); 
and (ii) 
we consider substances only in their \textit{causal role} on the generation of a concept, a  role that is constrained within the events during which a substance is perceived. We call such events, \textit{encounters} and (iii) we qualify this causal role in terms of two properties that substances have, as introduced below, and that we call \textit{Space Persistency} and \textit{Time Persistency}. Notice however that both \textit{Space Persistency} and \textit{Time Persistency}, as all the definitions provided in this paper, are given as properties of substance concepts. 

We assume that encounters are represented as \textit{spatio-temporal worms}, i.e., temporal sequences of \textit{frames}, where $\foS{i}{S}$ is a frame  for a substance $S$, each frame being encoded via a set of \textit{low-level visual features}.\footnote{Notationally, we use superscripts to mean elements of a sequence, and (optionally) the subscript $S$, to mean elements obtained from one or more encounters $\encS$ with the substance $S$, as in $\foS{i}{S}$, $\voS{i}{S}$, and $\objSii{i}{S}$. Different subscripts mean elements generated in possibly different sets of encounters. We omit the subscript whenever the substance we are referring to is clear from the context.}
We represent encounters, by exploiting the \textit{Space Persistency} of substances, namely the fact that, in time, substances change very slowly their spatial position. 
Because of space persistency, during an encounter, any two adjacent frames will be very similar, while this will not necessarily be the case with two non adjacent frames.
%
We model Space Persistency in terms of \textit{Frame Similarity (Dissimilarity)}, written
$\foSi{S_1}\simeq \foSi{S_2}$ ($\foSi{S_1}\not \simeq \foSi{S_2}$).
Given Frame Similarity, we define \textit{visual objects},  where $\voSi{S}$ is a visual object for a substance $S$, as sequences of adjacent frames where the last
frame is \textit{similar} to the first, and \textit{encounters} $\encS$ 
as sets of visual objects, i.e.,:
\begin{equation}
\label{eq:encounter}
\encS = \{ \voS{1}{S},\dots,\voS{n}{S} \}.
\end{equation}

\begin{figure}[t]
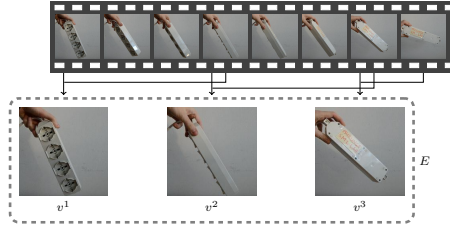

    \centering
    \includestandalone[width=.5\textwidth]{hyp2}
    \caption{\label{fig:visual_objects}Example of an encounter. For better visualization, each
      visual object is represented, here and later, as its first frame.}
      
\vspace{-0,4cm}
\end{figure}

\begin{figure}[h]
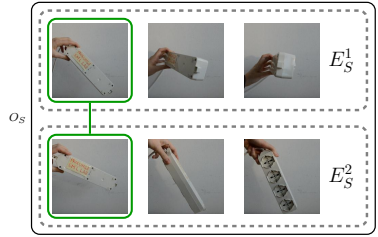

    \centering
        \includestandalone[width=.4\textwidth]{hyp1} 
        \caption{\label{fig:genus} A single object consisting of two
          encounters. The green line connects two similar visual
          objects.}
\end{figure}

\noindent Figure~\ref{fig:visual_objects} reports an example of an encounter consisting of eight frames organized in three visual objects. Notice how having multiple similar frames in the same visual object makes it quite robust to local contextual variations.
The first time a substance $S$ is perceived as a new \textit{object}, that object will consist of a single encounter; but this object will be enriched by subsequent encounters. 
We model this situation by taking objects to be the set of all the different visual objects collected by the different encounters. Let $\EoS{1}{S}, ..., \EoS{m}{S}$ be a set of encounters. Then we have:
\begin{equation}
\label{eq:objdef}
\objS = \cup_i\ \EoS{i}{S} =
\{ \voS{1}{S},\dots,\voS{n}{S} \} = 
\{ \voS{i}{S} \}.
\end{equation}
\noindent
This situation is well
represented in Figure~\ref{fig:genus} where each row is a  different encounter.

\section{Building Substance concepts}
\label{sec3-createsc}

Objects as substance concepts get cumulatively built in time.
Let $\EoS{1}{S}, ..., \EoS{m}{S}$ be a sequence of encounters. Let $\objS$ be an object defined as in Equation (\ref{eq:objdef}). Then $\objS$ is incrementally constructed as follows:
\begin{align}
\label{eq-buildobj}
1.\ \ \ & 
\textsc{AddObject}(\calM, \EoS{1}{S})  \\ 
2.\ \ \  & 
\textsc{updateObject}(\calM, \objSii{i}{S}, \objSii{i-1}{S} \cup \EoS{i}{S}),&  i=2,... \nonumber \\ 
3.\ \ \  & \objSi{S}\ \textsc{is}\ \objSii{i}{S},& i=1,...\nonumber
\end{align}

\noindent where: 
\begin{itemize}
  
    \item \textsc{AddObject} creates a new object $\objSii{1}{S}$ in the \textit{cumulative memory} $\calM$ of the objects perceived so far;  
    \item $\objSii{i}{S}$ is an object as perceived after any given number $i$ of encounters;  and
    \item $\objSii{i-1}{S} \cup \EoS{i}{S} \in \calM$ is the  cumulative memory of $\objSii{i}{S}$;
    \item \textsc{updateObject} updates the current memory $\objSii{i-1}{S}$ of an object with the visual objects coming from $\EoS{i}{S}$, thus constructing $\objSii{i}{S}$;
    
    \item The construct \textsc{is} in Item 3 is the formal statement assessing that we take objects as the cumulative memory of what has been perceived so far.
\end{itemize}
\noindent
A first observation is that item 3 implicitly states that  substance concepts evolve in time, i.e., that they are  perdurants. In this perspective, $\objSii{i-1}{S}$, $\objSii{i}{S}$, $\EoS{i}{S}$ and also $\objSi{S}$, are all partial views of $S$, all contributing to the construction of $\objSi{S}$. This process of object construction may eventually terminate if the appearance of an object does not change. However an object may also keep evolving. Thus, for instance, the current encounter with \textit{Frank} may contain visual objects which are quite dissimilar from the ones encountered earlier on, for instance because of the different age  (e.g., fifteen vs. thirty-five). 

A second observation is that the process described in Equation (\ref{eq-buildobj}), and in particular the decision of which between step 1 and step 2 must be applied, depends on the ability to recognize whether the current encounter is a partial view of an object already recognized. But how to decide? 
Let us write $\objSi{S_1} = \objSi{S_2}$ to mean \textit{Object Identity}, namely that the two substance concepts are two (partial) views of the \textit{same} object, rather than two views of two \textit{different} objects. This may, in fact, be the result of two different sequences of encounters with the same object. 
Let us also write
$\objSi{S_1} \not = \objSi{S_2}$ to mean \textit{Object Diversity}.  Then, Item 2 will be applied only for that object $\objSi{S}$ such that
 $\objSii{i-1}{S} = \EoS{i}{S}$, while Item 1 will be applied whenever $\objSii{i-1}{S} \not = \EoS{i}{S}$ for all objects in $\calM$. 
 
 The complications arising in the decision on Object Identity depend on two main factors.
 The first is that the \textit{correlation between substances and substance concepts is many-to-many}.\footnote{This corrects the wrong statement, made in \cite{giunchiglia2016fois}, that there is one-to-one mapping between substances and substance concepts} To reiterate an example from the previous section, the same substance can be perceived as \textit{Karl}, as a \textit{man} or as a \textit{person} while, vice versa, the same substance concept, e.g., \textit{man} can be recognized from multiple individuals. In other words, we need to decide at which level, in the visual subsumption hierarchy, the current encounter for the same substance should be assigned.
 The second issue is that, independently of the level of the subsumption hierarchy, the decision on Object Identity must made taking objects to be endurants, as represented by classification concepts, being classification concepts what is used by humans in their everyday interaction and classification activities.
  Object Identity is a much richer notion than visual similarity as it involves considerations like language, culture, function of the objects, and much more, see, e.g., \cite{hirsch1,noonan1, guarino1999role}. Among other things notice how we have $\objSi{S}= \objSii{i}{S},\ i= 1,...$, this meaning that Object Identity is invariant in time.
 As a consequence, there is a \textit{many-to-many correspondence between substance concepts and classification concepts}, as also extensively exemplified in \cite{giunchiglia2016fois}.
 
The double many-to-many mapping from substances to substance concepts and then from substance concepts to classification concepts is the main cause of the inherent ambiguity which appears in the identification of objects. This phenomenon is well known in Computational Linguistics and it is the cause of the so-called \textit{lexical gaps}, namely concepts which are lexicalized in one language but not in other languages \cite{giunchiglia2018one}. Things are made even worse when, even within the same language, one considers the subjective behaviour of individuals; see the two examples in Section \ref{sec-lsm}. Notice that the problem is not that of constructing a hierarchy of meanings; in Section \ref{sec-visual} we show how this can be done based on the visual similarity of objects as defined as in Equation (\ref{eq:objdef}). The problem is that such a hierarchy  will almost inevitably suffer from the SGP and, therefore, will not achieve the goal of aligning classification concepts and substance concepts. 
The solution we propose is articulated in the following main assumptions:

\begin{enumerate}
    \item We assume that the fact that two objects are visually similar is a \textit{necessary condition} for object identity. This assumption is well grounded in our everyday experience and also made in the mainstream Computer Vision research. To this extent, we introduce the notions of \textit{Visual Object Similarity (Dissimilarity)}, written $\voSi{S_1}\simeq \voSi{S_2}$ ($\voSi{S_1}\not \simeq \voSi{S_2}$) and of \textit{Object Similarity (Dissimilarity)}, written $\objSi{S_1}\simeq \objSi{S_2}$ ($\objSi{S_1}\not \simeq \objSi{S_2}$). Notice that we need to define what visual similarity is, given that, as discussed above, the same object can appear in many different ways; this will be discussed in Section \ref{sec-algorithm}. 
    \item We assume, as also 
implicit in Millikan's quote, that substances have a property of \textit{Time Persistency}, namely some form of time invariance in how they appear across encounters. This assumption allows us to  compare, up to a point, visual objects coming from different encounters. Notice that how space and time persistency operate is specific to the objects being considered, no matter whether instances or concepts. Thus, for instance, \textit{Karl} will keep having white hair while \textit{Frank} will keep having black hair and mustaches. Analogously, humans, like all animal species, 
are characterized by a \textit{homeostatic mechanism} which causes them to possess a certain
set of common traits (e.g., their shape, how they move)
that often, but not always, make them look similar \cite{giunchiglia2016fois}. 
The key consideration here is that, once an observer has subjectively decided what is the object that she is trying to recognize from a substance $S$, the criteria for object identity do not change. In other words, time persistency applies not only to the perceived object but also to the perceiving subject.
\item We organize objects in a visual subsumption hierarchy, exactly like the one used in lexical semantics, but with the key difference that \genus and \differentia are computed in terms of the substance concepts' visual properties, as represented by the visual objects. This allows to deal with the problem of the many-to-many mapping between substances and substance concepts.
\item 
\textit{Last, but not least}, we deal with the many-to-many mapping between substance concepts and classification concepts by relying on the key role of the \textit{user supervision}.
 This transformation is crucial to the integration of human vision, where objects keep evolving in time (i.e., they are perdurants), and language-based reasoning, which thinks of objects as being completely described in any moment in time (i.e., they are endurants).  
 \genus and \differentia can be computed in a completely unsupervised manner, via object similarity. However, the user feedback, which is given \textit{only} on \genus and \differentia, guarantees that the machine-built hierarchies largely coincide, modulo recognition mistakes, with the hierarchies that a user would build. Notice how this supervision is unavoidable, that it is exactly the same type of supervision that a mother would give to her child, and that it is subjective, evidence being also that different languages conceptualize different objects \cite{giunchiglia2018one}. 
\end{enumerate}

As a last remark, notice that in the visual hierarchy mentioned in item 3, all nodes are labeled only by substance concepts. Instead, in lexical semantics hierarchies, nodes are labeled by (classification) concepts, e.g., \textit{man}, and instances e.g., \textit{Frank}. In other words, as correctly pointed out by R. Millikan \cite{millikan2000clear}, but see also \cite{giunchiglia2016fois}, from a perception point of view, the usual KR distinction between concepts (usually modelled as sets of instances) and instances does not apply.

\section{Object Subsumption and Identity}
\label{sec-visual}

As from Equation  (\ref{sec-lsm}), objects, represented as substance concepts, are sets of visual objects. The idea is to  exploit this fact to build a hierarchy of objects based on visual similarity. As from the discussion at the end of the previous section, this hierarchy gives us only the necessary conditions for object identity. In the following we assume that the two hierarchies of substance concepts and classification concepts coincide assuming that the user feedback is used to validate the choices made. The algorithm in Section \ref{sec-algorithm} will show how this is done in practice by suitably asking feedback to the user.

As from Section \ref{sec-lsm}, a lexical semantics hierarchy can be seen as the iteration of many depth $2$ hierarchies, each with its own \genus and \differentia. Therefore, without lack of generality, in the following we focus on hierarchies of depth $2$. The main goal below is to restate the conditions for \genus and \differentia, informally stated in Section \ref{sec-lsm} for classification concepts, in terms of formally defined conditions on substance concepts.
Let us assume that we are given a genus object \genusobj. In the general case the construction of \genusobj will happen recursively from the top node, i.e., \textit{thing}. Then, let us define $\same(\objSi{S_1}, \objSi{S_2})$ and $\different(\objSi{S_1}, \objSi{S_2})$ as two binary boolean functions which discriminate over objects, based on their 
visual objects. We enforce the four roles in Section \ref{sec-lsm} by enforcing the following three constraints:
\begin{equation}
\label{eq-Genus}
 \neg\same(\objSi{S_1}, \objSi{S_2})\longrightarrow \objSi{S_1} \not = \objSi{S_2},  
\end{equation}
\begin{eqnarray}
\label{eq-Differentia}
\begin{tabular}{c}
 $\same(\objSi{S_1}, \objSi{S_2})  \longrightarrow$\\
$(\objSi{S_1} = \objSi{S_2}  \longleftrightarrow \neg\different(\objSi{S_1}, \objSi{S_2})),$ 
\end{tabular}
\end{eqnarray}
\begin{equation}
\label{eq:genusprop}
\objSi{S_G} \subseteq \objSi{S_1} \cap \objSi{S_2}.
\end{equation}
where  $\objSi{S_G}$ is the \genusobj of $\objSi{S_1}$, $\objSi{S_2}$. Notice how the specifics of  \genus and \differentia are left open, we only require that they are both computed out of the visual objects in input, i.e., $\objSi{S_1}$, $\objSi{S_2}$ and that they satisfy the three constraints above. This is on purpose as it gives us freedom in many dimensions, e.g., of the specifics of the learning algorithms used, of how visual similarity and/or object identity are defined, and also of how \same and \different are defined in any different layer of the hierarchy under construction. The algorithm in Section \ref{sec-algorithm} will instantiate the missing information selecting, for each decision point, one among the many possible options.

Let us concentrate on the constraints. They satisfy the following intuitions.  \textit{First}, they satisfy the four criteria defined in Section \ref{sec-lsm}.  Equation (\ref{eq-Genus}) formalizes \textit{Role 1} and \textit{Role 2} of \genus while Equation (\ref{eq-Differentia}) formalizes \textit{Role 1} of \differentia. Equation (\ref{eq:genusprop}) formalizes \textit{Role 2} of \differentia; in fact from
Equation (\ref{eq:genusprop}) we have $\objSi{S_G} \subseteq \objSi{S_1}$. To have $\objSi{S_G} \not = \objSi{S_1}$, $\objSi{S_1}$ must have at least a visual object $\voS{i}{S_1}\not\in \objSi{S_G}$. Then there are two cases, either $\voS{i}{S}$ is such that \different holds, in which case we are done (from Equation (\ref{eq-Differentia})), or this is not the case, in which case 
 $\objSi{S_G} = \objSi{S_1}$, namely that visual object is irrelevant to the identity of $\objSi{S_1}$. Notice how this latter case does not rise if we take \genusobj to be exactly the intersection. Equation (\ref{eq:genusprop}) captures the
intuition that the visual objects which are not considered belong to both objects by chance. Thus for instance, \textit{Karl} and \textit{Frank} might happen to have had, when observed, a red sweater. But red sweaters are not a characteristic of \textit{men}. 
\textit{Second}, Equation (\ref{eq-Genus}) captures the fact that \same provides necessary but not sufficient conditions for object identity. \textit{Third}, Equation (\ref{eq-Differentia}) provides necessary and sufficient conditions for two objects to be different, but under the assumption that \same holds. Namely, \different can be applied only after having discarded all the objects which do no satisfy \same. 

The three constraints above allow us to build the desired subsumption hierarchy.
 Let us write $\objSi{S_j} \sqsubseteq \objSi{S_i}$ ($\objSi{S_i}\sqsupseteq \objSi{S_j}$)  and say that 
  $\objSi{S_j}$ \textit{is subsumed by} $\objSi{S_i}$
  ($\objSi{S_i}$ \textit{subsumes} $\objSi{S_j}$) to mean that the visual objects of  $\objSi{S_j}$ are a subset of those visual objects of $\objSi{S_i}$ which are relevant for the computation of \different (see discussion above on Equation (\ref{eq:genusprop})). We also write $\objSi{S_j} \sqsubset \objSi{S_i}$  and talk of \textit{strict} subset and subsumption to mean 
  $\objSi{S_j} \sqsubseteq \objSi{S_i}$ and $\objSi{S_j} \not = \objSi{S_i}$, and similarly for $\objSi{S_i}\sqsupset \objSi{S_j}$.
 Let us assume that  $\objSi{S_2}$ and  $\objSi{S_2}$ have the same \genusobj,
 $\objSi{S_G}$ namely, that
 $\same(\objSi{S_1}, \objSi{S_2})$ and, therefore, $\same(\objSi{S_G}, \objSi{S_2})$, $\same(\objSi{S_G}, \objSi{S_1})$ hold. Clearly, $\objSi{S_G} \sqsubseteq \objSi{S_1}$ and $\objSi{S_G} \sqsubseteq \objSi{S_2}$. 
 This makes the premise and therefore the consequence of  Equation (\ref{eq-Differentia}) hold of all three objects. We have the following cases (for compactness, below we write $\D$ to mean \different):

\begin{enumerate}
\item  $\D(\objSi{S_1}$, $\objSi{S_2})$, $\D(\objSi{S_1},\objSi{S_G})$ and $\D(\objSi{S_2}$, $\objSi{S_G})$: we have $\objSi{S_G} \sqsubset \objSi{S_1}$ and $\objSi{S_G} \sqsubset \objSi{S_2}$, namely the situation where all three objects are different;
\item  $\D(\objSi{S_1}$, $\objSi{S_2})$, $\neg\D(\objSi{S_1}, \objSi{S_G})$ and
  $\D(\objSi{S_2}$, $\objSi{S_G}$): we have
  $\objSi{S_G} = \objSi{S_1}$ and
   $\objSi{S_G} \sqsubset \objSi{S_2}$; 

\item  $\D(\objSi{S_1}$, $\objSi{S_2})$, $\D(\objSi{S_1}, \objSi{S_G})$ and $\neg\D(\objSi{S_2}$, $\objSi{S_G})$: we have
  $\objSi{S_G} = \objSi{S_2}$ and
   $\objSi{S_G} \sqsubset \objSi{S_1}$; 

\item  $\neg\D(\objSi{S_1},\objSi{S_2})$, $\D(\objSi{S_1}, \objSi{S_G})$: we have  $\objSi{S_G} \sqsubset \objSi{S_1}$ with $\objSi{S_1} = \objSi{S_2}$;
  
\item  $\neg\D(\objSi{S_1} \objSi{S_2})$, $\neg\D(\objSi{S_1}, \objSi{S_G})$: we have $\objSi{S_1} = \objSi{S_2} = \objSi{S_G}$.

\end{enumerate}
\noindent
 Two observations.  The first is that, under the assumption that  $\objSi{S_1}$ and  $\objSi{S_2}$ have the same \genusobj  $\objSi{S_G}$,  \same and \different provide us with \textit{necessary} and \textit{sufficient} conditions for both \textit{object identity} and \textit{object subsumption} and, therefore, they provide us with the means for building the depth $2$ sub-hierarchy under consideration. In fact as from the (only if) directions of clauses 2,3,4,5, two objects are the same if they have the same \genus and \different does not hold of them. Thus, taking into account the necessary conditions provided by Equation (\ref{eq-Genus}) we have:
\begin{eqnarray}
\label{eq:objid}
\objSi{S_1} = \objSi{S_2} & \longleftrightarrow & \same(\objSi{S_1},\objSi{S_2}) \quad \land \\ \nonumber
& & \lnot \different(\objSi{S_1},\objSi{S_2})
\end{eqnarray}
Furthermore, as from clauses 1,2,3,4, we have that \genusobj is the parent node of the objects of which it is the \genusobj, namely:
\begin{eqnarray}
\label{eq:subsumes1}
 \objSi{S_1} \sqsubset \objSi{S_G} & \longleftrightarrow & 
 \different(\objSi{S_1},\objSi{S_G}) 
\end{eqnarray}
\noindent
The concluding remark is that, so far, we have only dealt with hierarchies of depth two, but the reasoning above can be replicated to build hierarchies of any depth. Let us assume that we have a new object $\objSi{S_3}$ with  $\neg\same(\objSi{S_3}, \objSi{S_1})$ and, thus, $\objSi{S_1} \not = \objSi{S_3}$, $\objSi{S_2} \not = \objSi{S_3}$,  and $\objSi{S_G} \not = \objSi{S_3}$. At the same time, $\objSi{S_3}$ can share some visual objects with $\objSi{S_1}$ or $\objSi{S_2}$  which make \different false. Thus, for instance, a \textit{plane} is not a \textit{bird}, but they both \textit{fly}. Given, any two objects there is \textit{always} a \genusobj,  also when these objects are very different, and this is the key fact which allows for the construction of subsumption hierarchies of any depth. Notice how we may end up with a \genusobj which is the empty set, this being the limit case where the \genusobj is \textit{thing}: a generic object is something which has been perceived but with no associated visual objects.

\section{The learning algorithm}
\label{sec-algorithm}

We first provide a computational interpretation of the definitions introduced above and then we introduce the algorithm,
which should be seen as a first prototype and representative of a wide class of algorithms. Any algorithm will do as long as it satisfies the constraints for \genus, \differentia and the \genusobj. 
Let us analyze the definitions one by one.

\vspace{0.1cm}
\noindent
\textit{Frames.} We encode frames using an \textit{unsupervised} deep
neural network~\cite{caron2019unsupervised}, trained to perform a
combination of self-supervised and clustering tasks. Using an
unsupervised network allows to produce embeddings which are not
explicitly biased towards classes of objects, while, at the same time,
complying to the assumption that machines extract features from what
they perceive, autonomously, as humans do.  We define \textit{frame
  similarity} as the Euclidean distance between frame encodings.

\vspace{0.1cm}
\noindent
\textit{Visual objects.} We define them as contiguous sequences of frames, and we represent them as the average between the frame encodings. We assume for robustness that visual objects are of a fixed limited length.
Visual object are perceived by a procedure, named \textsc{perceive}, which returns an encounter as a set of visual objects, as from Equation (\ref{eq:encounter}). We model
\textit{visual object similarity} as a diversity threshold on the distance between visual objects:
\begin{equation}
    \label{eq:vsim}
    \voS{i}{}\simeq \voS{j}{} \defas d(\voS{i}{}, \voS{j}{}) < \theta
\end{equation}

\noindent
\textit{Objects.} We define objects as from Equation (\ref{eq:objdef}), i.e., as sets of visual objects
extracted from sets of encounters. 
We define 
{\em object similarity}, analogously to visual object
similarity, as a diversity threshold on the distance between
objects:
\begin{equation}
    \label{eq:objsim}
        \objSi{S_1} \simeq \objSi{S_2} \defas d(\objSi{S_1},\objSi{S_2}) < \theta
\end{equation}
where the distance between objects is defined
as the minimal distance between their respective visual objects:
\begin{equation}
    \label{eq:objdist}
        d(\objSi{S_1},\objSi{S_2}) = \min_{\voS{i}{} \in \objSi{S_1}}\min_{\voS{j}{} \in \objSi{S_2}} d(\voS{i}{},\voS{j}{})
\end{equation}
By keeping the same threshold as for visual object similarity, we have that object similarity holds when two objects have at least two similar visual objects:
\begin{equation}
    \label{eq:objsim}
        \objSi{S_1} \simeq \objSi{S_2} \iff \exists \voS{i}{} \in \objSi{S_1}, \exists \voS{j}{} \in \objSi{S_2}\;: \;
 \voS{i}{}\simeq\voS{j}{}.
\end{equation}

\vspace{0.1cm}
\noindent
\textit{Genus.} We implement \same as a Boolean function which computes object similarity:
\begin{eqnarray}
  \label{eq:genus}
\same(\objSi{S_1},\objSi{S_2}) \defas \objSi{S_1} \simeq \objSi{S_2}
\end{eqnarray}
In other words, we take object similarity to be a sufficient condition for \same to hold. This is appropriate for the base case of hierarchies of depth two, with the implicit assumption that objects with different genus are all instances of a generic {\it thing} object. We leave the generalization to deeper hierarchies to future work. A hint on how to perform such generalization can be found in the Conclusion Section.

\vspace{0.1cm}
\begin{figure}[t]
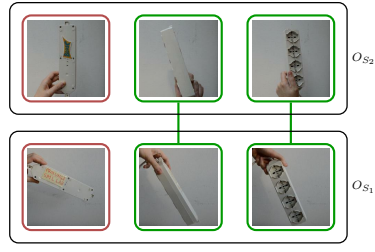

    \centering
        \includestandalone[width=.4\textwidth]{hyp3} 
        \caption{\label{fig:differentia} Two distinct objects which
          share the same \genusobj. The green lines connect similar visual
          objects, while the visual objects representing the \differentia are
          highlighted in red.}
\end{figure}

\noindent
\textit{Differentia.} We implement \different as a boolean function which holds for two objects with the same \genusobj 
if there is no visual object, aside the ones in their \genusobj, which makes the two objects similar, namely:

\begin{eqnarray}
\label{eq:differentia}
\different(\objSi{S_1},\objSi{S_2}) & \defas &  \nexists \voS{i}{} \in \objSi{S_1} \setminus \genusfun(\objSi{S_1}) \nonumber. \\
& & \nexists \voS{j}{} \in \objSi{S_2} \setminus \genusfun(\objSi{S_2}) \nonumber: \\
& & \voS{i}{} \simeq \voS{j}{} \quad
\end{eqnarray}
\noindent 
where \genusfun is a function which computes the \genusobj, as:

\begin{eqnarray}
\label{eq:genusof}
    \genusfun(\objSi{S_1}) & \defas & \{\voS{i}{} \in \objSi{S_1} |  \exists \objSi{S_2}, \exists  \voS{j}{} \in \objSi{S_2} :  \textsc{sG}(\objSi{S_1},\objSi{S_2}) \land \voS{i}{} \simeq \voS{j}{}\}
\end{eqnarray}
\noindent where the function $\textsc{sG}(\objSi{S_1},\objSi{S_2})$ returns true if the user
in the past gave supervision (see below), telling the algorithm that $\objSi{S_1}$ and $\objSi{S_2}$
share a \genusobj while being different.
Figure~\ref{fig:differentia} shows two objects with the same \same (green lines) but for which also \different holds (the two red visual objects are different). 

\vspace{0.1cm}
\noindent
\textit{User feedback.} We use two functions $\textsc{askSameGenus}$ and $\textsc{askDifferent}$ which 
ask the user, when available, about \same and \different between an encounter $E$ and an object $O$ stored in memory. If the user is not available, they return $\same(\objnn,\enc)$ and $\different(\objnn,\enc)$,
respectively. Notice how the user intervention is exploited \textit{exactly and only} in the computation of \same and \different, in order to
consolidate object similarity into object identity, as from the previous section.
The user feedback, collected by $\textsc{askSameGenus}$ and $\textsc{askDifferent}$, is exploited by a function
$\textsc{updateSimilarity}(\calM)$ whose goal is to adjust the diversity threshold $\theta$, see Equation (\ref{eq:vsim}), based on the knowledge available so far.
The threshold is computed using the strategy, proposed in \cite{erculiani2019continual}, each time a new supervision is provided by the user.
These supervisions are stored as a set:
$$\calK = \{ \langle \delta_i, y_i \rangle \mid 1 < i < |\calK| \}$$
where $\delta_i = d(O_i, E_i)$ is the distance between object-encounter pairs, coupled
with a boolean value $y_i = \textsc{askSameGenus}(O_i,E_i)$ containing the supervision of the user. The value of $\theta$
is computed solving the following  problem:
\begin{equation}
  \label{eq:lambda_r}
  \theta = \argmax_{\lambda}  \sum_{i=1}^{|\calK|} \mathbbm{1}(  (\delta_i < \lambda) \oplus \neg y_i )  
\end{equation}
\noindent where $\mathbbm{1}$ is the indicator function mapping $True$
to 1 and $False$ to 0, and $\oplus$ is the exclusive OR. \cite{erculiani2019continual} provides a strategy for how to efficiently solve this problem by performing a number 
of evaluations equal to $|\calK|$ .

\begin{algorithm}
    \caption{Build subsumption hierarchy}
        \raggedright
    \begin{algorithmic}[1]
    \Procedure{BuildSubsumptionHierarchy}{}
        \State $\calM \gets \emptyset$;
        \While{True}
            \State $\enc \gets \textsc{perceive}()$
            \State $\objnn \gets \textsc{getMostSimilarObject}(\enc, \calM) $
            \If {$\textsc{askSameGenus}(\objnn,\enc)$}
              \If {$\textsc{askDifferent}(\objnn,\enc)$}
                \State $\textsc{AddObject}(\calM, \enc)$
              \Else
                \State $\textsc{updateObject}(\calM, \objnn, \objnn \cup \enc)$
              \EndIf
            \Else
                \State $\textsc{AddObject}(\calM, \enc)$
            \EndIf
            \State $\textsc{updateSimilarity}(\calM)$
        \EndWhile
    \EndProcedure
    \end{algorithmic}
    \label{alg:hierarchy}
\end{algorithm}

\noindent
The algorithm building the subsumption hierarchy is implemented as the
infinite loop shown in Algorithm \ref{alg:hierarchy}. 
This algorithm is a direct implementation of the recursive construction of objects given in Equation (\ref{eq-buildobj}) via the test for object equality and subsumption as from Equations (\ref{eq:objid}), (\ref{eq:subsumes1}).
We use a function \textsc{getMostSimilarObject} 
which, given an object and a cumulative memory $\calM$ of all the objects perceived so far, returns the object which is most similar. The implementation of this function is based on the consideration that there are two possible cases. In the first, that same object was previously seen and, therefore, this is the object to be selected. In the second, the object was not previously seen, in which case there may be no objects sharing visual objects (no objects with the same \genusobj) or there may be one or more similar objects, possibly including the \genusobj, which share the \genusobj with the new object. 
Based on this intuition \textsc{getMostSimilarObject} returns the nearest already seen instance that satisfies the similarity constraint of Equation (\ref{eq:objsim}), if existing, otherwise it returns the most similar \genusobj, computed as described above.
Notice that we make the further simplifying assumption to ask the user, via $\textsc{askSameGenus}$, only for the most similar element. Thus the model is not guaranteed to keep a hierarchy always in line with the desires of the user. Ideally one should ask for supervision for every similar object. This choice was made to
limit the effort required to the user. We will deal with this problem in further research following the line of thought already started in \cite{erculiani2019continual}.

For what concerns lines 6--12 of the algorithm, we have the following: (i) in Line 8, it creates a new object because \different holds
for an object with the same \genus (as from Equation (\ref{eq:subsumes1}), this is the case when subsumption holds); (ii) in Line 10, it extends an already existing object for which \same holds but \different does not (as from Equation (\ref{eq:objid}), this is the case when we have object identity); in Line 12, it creates a new object corresponding to an instance of a new \genusobj (as from Equation (\ref{eq-Genus})). It is easy to see how the hierarchy satisfies, for any given sequence of encounters, the five conditions provided in the previous section.

\hl{Each iteration of the algorithm requires a forward step in the
  embedding network followed by a nearest-neighbour search in the
  memory of stored encounters. A simple linear search is sufficient
  for real-time interaction for reasonable sized datasets, compatible
  with the need for supervision by a single person. On the other hand,
  the approach can easily scale to arbitrary datasets by leveraging
  techniques like
  tree-decomposition}~\cite{Clarkson06nearest-neighborsearching}\hl{ or
  locality-sensitive-hashingh}~\cite{lsh2014,hash2017}\hl{ for efficient
  exact and approximate nearest-neighbor search. These approaches can
  be complemented with prototype selection
  strategies}~\cite{Bien_2011,garcia2012,Zhang2019}\hl{ to combine time and
  memory efficiency.}

As a concluding remark, notice how the algorithm satisfies the
requirements listed in the introduction: (i) the hierarchy is built
autonomously and it becomes a hierarchy of classification concepts
thanks to the user feedback, (ii) no assumption is made about the
input objects, (iii) the algorithm learns objects never seen before
and (iv) it incrementally learns how to recognize objects starting
from no objects.

\section{Experiments}
\label{sec-evaluation}

\def\ecaiegocentric{\cite{erculiani2019continual}\xspace}
\def \ecaimodel{\textsc{Fullower}\xspace}

\def \ourmethod{\textsc{GenusDifferentia}\xspace}

\def\ambseq{ambiguous\xspace}
\def\discseq{discriminative\xspace}
\def\thrds{threshold\xspace}
\def\ambds{ambiguity\xspace}

\def\thrlearn{threshold learning\xspace}
\def\amblearn{ambiguity learning\xspace}

\def\visobj{visual objects extraction routine\xspace}

\def \ablmean{AblationMean\xspace}
\def \ablamb{AblationNoAmbiguity\xspace}

\hl{The algorithm was implemented in PyTorch and the implementation
  can be freely downloaded at}
\url{https://github.com/lucaerculiani/towards-visual-semantics}. \hl{In
  all experiments we have used a moving average of size fifty and
  stride fifteen to create the visual objects. The embedding network
  is a standard VGG-16 architecture}~{\cite{vgg}}\hl{ which was
  pre-trained on the YFCC100M dataset}~{\cite{YFCC100M}}\hl{ in an
  unsupervised fashion with the DeeperCluster algorithm
(see}~\cite{caron2019unsupervised}\hl{ for the details of how the network
was pre-trained).}

\subsection{\hl{Dataset}}

\begin{figure}
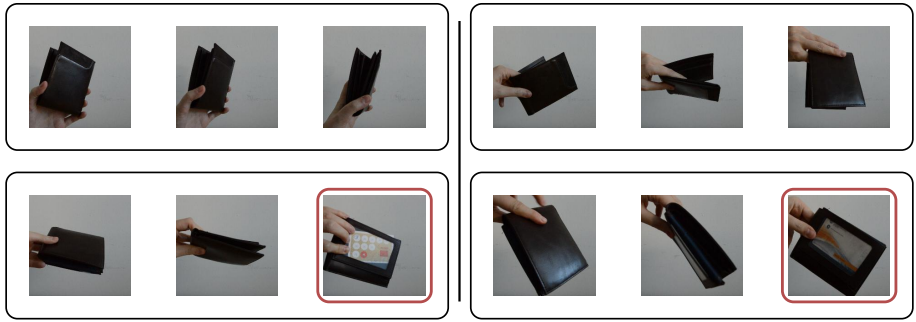

    \centering
        \includestandalone[width=.98\textwidth]{dataset} 
        \caption{\label{fig:dataset} \hl{Sample frames from the
          dataset. The left and right columns represent two wallets
          that only differ by the card they contain. The
          discriminative view is only present in the bottom row
          (highlighted in red).}}
\end{figure}

The main difficulty in setting up the experiments was that no existing
dataset matches the conditions we needed to properly evaluate our
framework. This setting requires a collections of objects that can be
grouped on the basis of their visual appearance. Inside each group,
all objects must have some partial views that make them
\emph{indistinguishable} from the other elements of the group (the
\genus), while at the same time having other views that enable the
discrimination of single objects (the \differentia). No public
dataset enforces this constraint. As a consequence we have created
our own data set, which consists of a collection of video sequences of
various objects, recorded while rotating or being deformed against a
blank background, making sure that each video contained only a {\em partial} 
view of the object. We have made the simplifying
assumptions of a blank background, which is clearly limiting for a
real world application. This assumption is motivated by the focus on
the recognition of \genus and \differentia, rather then on the
distinction between objects and background.
The resulting dataset\footnote{\hl{The dataset is freeely
    available at}
  \url{https://figshare.com/articles/dataset/small_re-identification_dataset_with_classes/14706003},
  \hl{where both raw data and precomputed embeddings can be
    downloaded.}} \hl{contains videos for five different types of
  objects: a coffee pod, a multiplug, a pencil case, a smartphone and
  a wallet.} For each object type we recorded videos for two different
  instances, that were different only for a certain view (as in Figure
\ref{fig:differentia}). For each object instance we recorded five
  videos that contain the discriminative view, and five that do
  not. \hl{Videos were recorded at 60 fps and lasted between 1 and 5
  seconds. Figure}~\ref{fig:dataset} \hl{shows some sample frames for
  wallets. The left and right columns represent two distinct wallets
  (that should be recognized as having the same \genus), that only
  differ by the card the contain. The top row shows (excerpts of)
  videos that do not contain the discriminative view (and should thus
  be predicted as not having \differentia), while the bottom row shows
  videos of the same objects where the differentia is visible (the red
  frame in each sequence).}

\subsection{\hl{Experimental results}}

In the following we report first qualitative and then quantitative
results in terms of capacity of the learning algorithm to recover the
notions of \genus and \differentia of the user. Below, we say that the
answer is correct when this is the case, incorrect otherwise.

\subsubsection{\hl{Qualitative results}}

\begin{figure}[]
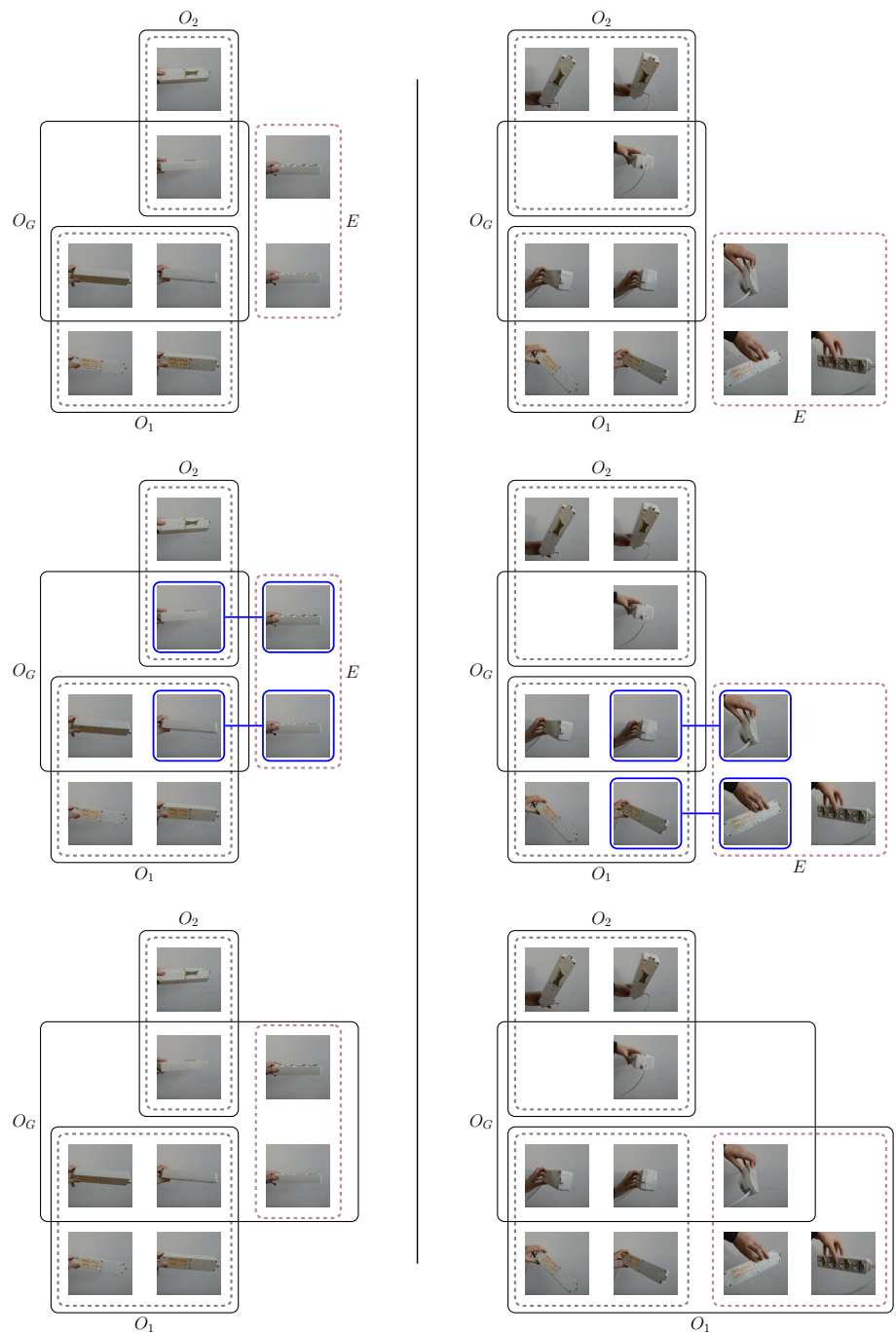

    \centering
        \includestandalone[width=.98\textwidth]{corretti} 
        \caption{\label{fig:corretti} Examples of two correct choices made 
        by the algorithm. 
        The left column depicts a case in which the machine correctly identified the \genus of the new encounter 
        (the encounter does not contain enough information for instance-level recognition). 
        The second column represents a case in which the new encounter was correctly identified as
        an already seen object. The new visual objects were added to the matched object. In addition, the \genusobj  
        was updated with the subset of visual objects matching it.}
\end{figure}

Figure~\ref{fig:corretti} shows two cases of encounters that were correctly processed by the algorithm with no user intervention. Each column represents the sequence of steps made to process a new encounter 
(the visual objects in the purple box), namely perception, recognition and memorization,
and the two columns represent cases giving rise to different choices made by the algorithm.
The encounters already present in memory are represented by gray dashed boxes, and 
the corresponding objects by black boxes. A box covering visual objects from multiple objects
represents the genus of that group of objects. The linked couples of 
blue visual objects represent items that were recognized as similar by the machine. In the left column,
a new encounter is correctly recognized as having the same \genus of two objects already stored in memory. The \genusobj is updated by incorporating the visual objects of the encounter. This is all the algorithm can do, as the encounter does not have enough visual information to allow for an instance-level recognition. In the right column, a new encounter is correctly recognized as being the same as an object stored in memory. The visual objects of the encounter are added to the retrieved object, while its \genusobj is updated with the visual objects that are found similar to it. Note how updating the object enriches its representation by including a viewpoint that was never observed before (the one showing the sockets). 

\begin{figure}
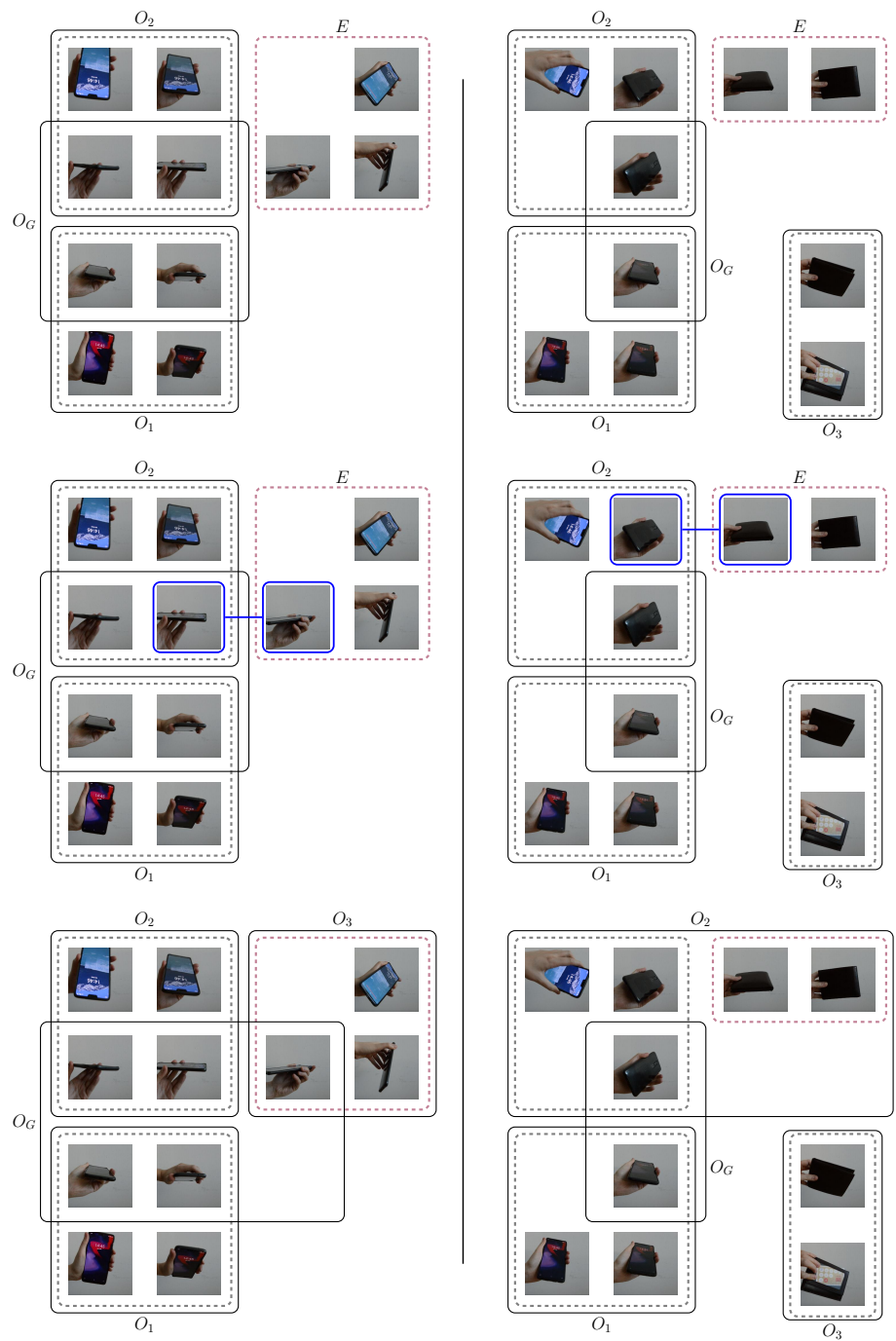

    \centering
        \includestandalone[width=.98\textwidth]{sbagliati} 
        \caption{\label{fig:sbagliati} Examples of two incorrect choices made 
        by the algorithm. 
        The left column depicts a case in which the machine recognized the correct \genusobj
        for the new encounter but not the correct instance. This led to the creation
        of a new separate object with that \genusobj.
        The second column represents a case in which the new encounter (containing a wallet) was mistakenly
        assigned to a completely different object (a smartphone).}
\end{figure}

Figure~\ref{fig:sbagliati} shows two cases of encounters in which the algorithm made choices which are not aligned with the user perspective. In the left column, the algorithm manages to recognise the \genus of the new encounter, but fails to realize that the encounter is actually the same as one of the objects it has seen before. In so doing it wrongly creates a new object in memory.
This error can be avoided if user feedback is available to answer an $\textsc{askDifferent}$ query (line 7 of Algorithm~\ref{alg:hierarchy}). The right column represents a case in which the new encounter was mistakenly identified as a completely different object. In this case, the availability of user supervision can prevent the algorithm from performing the wrong match (and spoiling the representation in memory of the retrieved object). The algorithm would in this case create a new object initialized with the encounter. Note however that in case the encounter was indeed an instance of an already stored object (but different from the one retrieved by the machine), or shared a \genus with it, asking feedback for the most similar object only as done in Algorithm~\ref{alg:hierarchy} would not suffice to discover it (see discussion in Section~\ref{sec-algorithm} on the limitations of this choice). Indeed, the algorithm should progressively ask for feedback on a sequence of objects (of decreasing similarity) until the user confirms the match, which could end up being too demanding for the user. A possible solution is that asking the user to provide names for objects and genuses, thus making the mapping between substance and classification concepts explicit. This however does not solve the problem entirely, as without full supervision the memory would contain objects without names. Note also that a purely name-driven supervision cannot work, for the very reasons that have been discussed when contrasting substance concepts with classification concepts (e.g., the user could provide the name of a genus when the new encounter also has a differentia). We plan to investigate in future work a hybrid similarity-driven and name-driven retrieval strategy in conjunction with the extension of the method to 
hierarchies of arbitrary depth.

\subsubsection{\hl{Quantitative results}}

\begin{figure}[h]
	\centering
	\begin{tabular}{c @{\hspace{0.0\tabcolsep}} c @{\hspace{0.0\tabcolsep}} c}
         & \enspace (a) & \enspace (b) \\[-0.5ex]
        \rotatebox[origin=l]{90}{ \scriptsize \sffamily \hspace{25pt} accuracy} & \includegraphics[width=.46\textwidth]{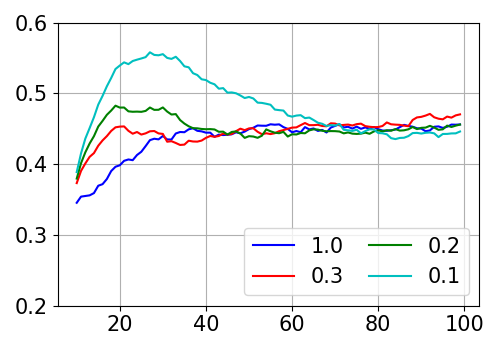} & \includegraphics[width=.46\textwidth]{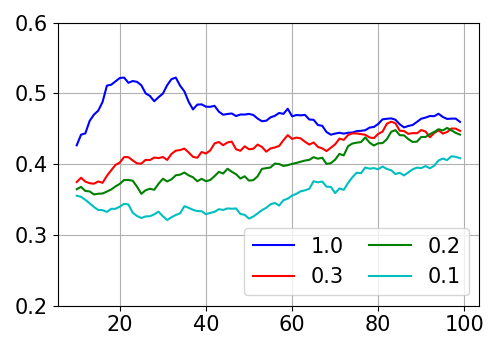} \\[-1.5ex]
         & { \scriptsize  \sffamily iteration} & { \scriptsize  \sffamily iteration}
	\end{tabular}
        \caption{\label{fig:results} Accuracy of prediction
        for \genus (a) and \differentia (b) respectively, for increasing number of iterations and different amounts of supervision (curves for different values of $\alpha$).}
\end{figure}

What described above provides a qualitative view of the behaviour of
the algorithm, which largely depends on the availability of user
feedback. We have also ran a quantitative evaluation showing how
recognition performance over time is affected by the amount of
supervision. The experiment is organized as follows. Sequences are
showed one after the other and at each iteration the user provides
supervision with probability $\alpha$.  We have run experiments for
different values of $\alpha$ with $\alpha \in \{1.0, 0.3, 0.2, 0.1\}$
and where $\alpha=1.0$ is the setting where supervision is always
available. In all settings the model is provided with a supervision
for each of the first five sequences in order to bootstrap the
estimation of the diversity threshold $\theta$.
We ask the model to predict  \same and \different at each iteration before receiving feedback from the user (if available, otherwise the algorithm prediction is used to update the memory). The results of the experiment are depicted in Figure \ref{fig:results}. 

Figure \ref{fig:results}(a) presents the accuracy computed for the prediction of \genus. A prediction is correct if the new encounter shares a \genus with an object in memory and the algorithm retrieves the correct \genusobj, or the algorithm correctly identifies the encounter as a completely novel object.
The plotted results are computed as the mean accuracy of the prediction 
over two thousand different runs, each with a different order of the sequences, 
smoothing the curves using a moving average of length five.
Surprisingly enough, in the first half of the experiment the smaller the 
supervision the better the accuracy. This is due to the fact 
that at the beginning, most sequences contain new objects, thus the more the supervision the higher the bias of the model to predict a new sequence as unseen. 
This bias progressively fades away proceeding with the experiment, and all models end up achieving similar results on average. \hl{These results suggest that even a very limited amount of supervision is sufficient to learn a reasonable value for the diversity threshold, which is what the algorithm needs in order to retrieve objects with the same genus if stored in memory (see Equations} ~\ref{eq:vsim} and \ref{eq:objsim}).

Figure \ref{fig:results}(b) shows the accuracy of the prediction of \different, 
over the subset of sequences for which the \genus is predicted correctly. 
\hl{The task here is more complex, as the algorithm needs to gather enough supervision to characterize the genus object (\genusobj) so as to identify the} \differentia \hl{(see Equations}~\ref{eq:differentia} and \ref{eq:genusof}). Indeed, in this case the greater the amount of supervision, the better the 
model is capable of recognizing whether the new sequence contains enough information to identify the correct instance. Apart for the setting with least supervision ($\alpha = 0.1$), for which the performance gap with respect to the fully supervised case stays rather large, the different models end up achieving comparable performance.

Overall, these preliminary results indicate that the algorithm is capable of progressively acquiring the notions of \genus and \differentia with reasonable accuracy despite seeing a small number of examples and receiving supervision on a fraction of them. 
\hl{These results are a proof-of-concept of the feasibility of the approach. The main limitation of this work is the fact that we are restricting ourselves to hierarchies of depth 2. Extending the learning algorithm to deeper hierarchies is indeed our main target for future research, as discussed in the conclusion of the paper. This will also require building a much bigger data set, having as main reference the  Imagenet dataset} \cite{ilsvrc2015}. \hl{Building this data set and the corresponding hierarchy of classification concepts is a research task in itself.}

\section{Related work}
\label{sec-related}


As already hinted in the introduction, this work is grounded in the work done in Teleosemantics. At the same time, the distinction between substance concepts and classification concepts resembles the two types 
of concepts proposed by R. Millikan and J. Fodor, see also their debate on Recognitional Concepts  \cite{1998-fodor, millikan1998more}. In fact, substance concepts map quite naturally to Millikan's recognitional concepts while classification concepts seem to be a good conceptualization of Fodor's work on the structure of semantic theories \cite{1963-fodor}. The work provided here suggests that we need both types of concepts, which functionally serve different problems, the crucial issue being how to keep them aligned.

This work constitutes a major shift from mainstream KR and Computer Vision in four dimensions. First, it treats objects as perdurants, where classification concepts are, instead, endurants.
In other words, we assume that we have two (very different) representations for anything we perceive, e.g., a \textit{person}. Objects are assumed to be represented only partially and to evolve in time building (modulo forgetting) richer and richer but never complete visual representations. 
Second, it uniformly models instances and classification concepts as substance concepts, and, therefore, as sets of visual properties. 
Thus, substance concepts, i.e., objects, are visual representations of both classification concepts and instances.
Third, object visual similarity is \textit{not} taken to be the same as object identity, this latter notion applying only to classification concepts. 
Fourth the user is never asked about the name of an object but only about \genus and \differentia.
 
 The work proposed in this paper is  a (first step towards the) solution of the SGP. The previous work so far has been on how to integrate feature-level information with semantic level information. Thus, some early work has proposed to encode semantic information via ontologies \cite{hare2006mind}, others propose to use tags or similar high-level features \cite{ma2010bridging,elahi2017exploring}, others propose to involve users using active learning \cite{tang2011semantic}, most recently it has been proposed that the semantic gap should be handled in DNNs when aggregating multi-level features \cite{pang2019}. The common denominator is that this work exploits classification concepts, rather than substance concepts, and that it does not build the subsumption hierarchy. All these proposals do not provide a general solution to the SGP. A fair amount of work has also been done trying to model objects
in a way which is compliant to how humans think about objects.
Most of this work, motivated by Robotic applications has concentrated on identifying the function of objects see, e.g., \cite{dimanzo1989understanding,stark1991achieving,bogoni1995interactive,woods1995learning,pechuk2005function}. But because of its very purpose, this work models objects as classification concepts.

The visual hierarchy proposed in this work naturally reminds of the work on hierarchies done in Computer vision and, in particular the work on  ImageNet~\cite{deng2009imagenet}. The introduction of the ImageNet dataset
and its associated challenge~\cite{ilsvrc2015} has boosted image classification towards (and even beyond) human-level performance. While most research has focused on fine-grained classification of (subsets of) leaf classes, hierarchical classification has also been directly addressed~\cite{yan2015}. However, this work assumes a {\em predefined} hierarchy given in advance, as well as a fixed set of examples to learn from. Furthermore, our focus is on the classification of videos of objects rather than static images. 
\hl{It is part of out future work to collect a dataset of object videos that resembles the ImageNet dataset in terms of size and depth of the hierarchy. A promising direction consists in leveraging the recent developments in terms of Embodied AI simulators}~\cite{duan2021survey}, \hl{that however need to be adapted in terms of quantity, diversity and granulatity of the concepts that can be represented.}

In recent years there has been a growing interest  towards open
world~\cite{bendale2015towards,rudd2018extreme}, continual and
lifelong learning~\cite{PARISI2019}. Most approaches focus on the
sequential learning of novel classes via class-specific training
sessions, trying to avoid catastrophic
forgetting~\cite{goodfellow2014empirical} by e.g., parameter
regularization~\cite{kirkpatrick2017overcoming,zenke2017continual},
model capacity expansion~\cite{yoon2018,rusu2016progressive} or
task rehearsal~\cite{shin2017continual,ven2018generative}. Alternatives accounting for unsupervised~\cite{Rao2019} or task agnostic~\cite{zeno2018task} settings have also been recently explored. However, the underlying assumption is always the presence of a predefined (possibly not explicit) set of classes, that are progressively presented to the algorithm. Even the open-world classification
setting~\cite{bendale2015towards,rudd2018extreme}, where the learner
should be able to tell if an entity does not belong to the
set of known classes (the so-called open-set
classification~\cite{scheirer2013toward,bendale2016towards}), requires
a specific training session in order to incorporate novel classes. On
the other hand, few-shot learning
methods~\cite{vinyals2016matching,santoro2016meta,ravi2017optimization},
that address the scarcity of data using similarity-based or meta
learning approaches, are typically closed-world and offline, with
well-separated training and testing phases. A fully online
incremental and agnostic setting where the hierarchy of objects emerges from the combination of encounters and feedback from the user is beyond the scope of these approaches.

\section{Conclusion}
\label{sec-conclusion}

In this paper we have provided the first steps towards a general theory of visual semantics. The  ultimate goal is to understand the general mechanisms by which it is possible to align the meaning of words with the perception of the objects named by those words. The main foundational contribution of this paper is the distinction between substance concepts and classification concepts, the first being modeled as perdurants the latter as endurants, and the mechanism by which these two different types of concepts must be aligned. This latter result has highlighted the crucial role of humans, not so much to tell machines what objects are, machine can learn this by themselves, but to make sure that what they learn is coherent with how humans perceive the world.

The future work will proceed in many directions. The first will be the extension of this work to hierarchies of any depth.
For what concerns the algorithm, the main requirement is the addition of a system to build a new \genusobj on top of an existing \genusobj. The general idea we foresee is that of ditching the global threshold mechanism in favor of a different similarity threshold (or metric) for each \genusobj. The second extension is that of combining similarity-based retrieval with name-based retrieval, in order to quickly but reliably identify the position(s) in the memorized hierarchy where to put the new encounter. We also plan to reduce the burden of the user by introducing online active learning strategies (see~\cite{erculiani2019continual} for an initial solution in a flat instance-level recognition task). 
%

%
Finally, the algorithm should be adapted to deal with tougher visual contexts including variable background, occlusions, noisy feedback etc.

\section*{Acknowledgements}

This paper was supported by the \emph{``WeNet - The Internet of Us"} and the {\em TAILOR} projects, funded by the European Union (EU) Horizon 2020 programme under GA number 823783 and 952215, respectively.

\section*{Conflict of Interest}

On behalf of all authors, the corresponding author states that there is no conflict of interest.

\bibliographystyle{splncs04}
\bibliography{ref}


\end{document}